\documentclass[10pt,twocolumn]{IEEEtran}
\usepackage{graphicx}           
\usepackage{amsfonts}
\usepackage{url}
\usepackage{times,epsfig}
\usepackage{subfigure}
\usepackage{multirow}
\usepackage[justification=centering]{caption}
\usepackage{fancyhdr}
\begin{document}

\pagenumbering{arabic}
\title{Incentive Design for Efficient  Federated Learning in Mobile Networks: A Contract Theory Approach}

\author{Jiawen Kang, Zehui Xiong, Dusit Niyato,~\emph{Fellow, IEEE}, Han Yu, Ying-Chang Liang~\emph{Fellow, IEEE} \\	and Dong In Kim, \emph{Fellow, IEEE}
\IEEEcompsocitemizethanks{
This work was supported in part by WASP/NTU M4082187 (4080), Singapore MOE Tier 1 under Grant 2017-T1-002-007 RG122/17, MOE Tier 2 under Grant MOE2014-T2-2-015 ARC4/15, NRF2015-NRF-ISF001-2277, EMA Energy Resilience under Grant NRF2017EWT-EP003-041. 

Jiawen Kang, Zehui Xiong, Dusit Niyato and Han Yu are with School of Computer Science and Engineering, Nanyang Technological University, Singapore. Emails: \{kavinkang, dniyato, han.yu\}@ntu.edu.sg and  zxiong002@e.ntu.edu.sg.  Ying-Chang Liang is with Center for Intelligent Networking and Communications, University of Electronic Science and Technology of China, China. Email: liangyc@ieee.org. Dong In Kim is with School of Information \& Communication Engineering, Sungkyunkwan University,  Korea. Email: dikim@skku.ac.kr.

Compared with the published version in IEEE Xplore digital library, we have made some typo corrections in this version. 
} \vspace*{-0mm}
}
\maketitle
\pagestyle{headings}

\begin{abstract}
To strengthen data privacy and security, federated learning as an emerging  machine learning technique is proposed to enable large-scale nodes, e.g.,  mobile devices,  to distributedly train and globally share models without revealing their local data. This technique can not only significantly improve privacy protection for mobile devices,  but also ensure good performance of the trained results collectively. Currently, most the existing studies focus on optimizing federated learning algorithms to improve model training performance. However, incentive mechanisms to motivate the mobile devices to join  model training have been largely overlooked. The mobile devices suffer from considerable overhead in terms of computation and communication  during the federated model training process. Without  well-designed incentive, self-interested mobile devices will be unwilling to join federated learning tasks, which hinders the  adoption of federated learning. 
To bridge this gap, in this paper, we adopt the contract theory to design an  effective incentive mechanism  for simulating the mobile devices with high-quality (e.g., high-accuracy)  data to participate in federated learning. Numerical results demonstrate that the proposed mechanism is  efficient for federated learning with improved learning accuracy.

\end{abstract}


\section{Introduction}
With the rapidly improving computation and communication capabilities of mobile devices,  many novel mobile applications based on machine learning techniques, e.g.,  Google Translate APP,  are emerging to bring excellent experience to mobile users  \cite{yang2019federated}. Although the machine learning techniques dramatically enhance the performance of  mobile applications,  traditional machine learning techniques require mobile devices to directly upload  user data with potentially sensitive private information to a central server for model training \cite{zhu2018multi}.  This causes not only large computation and storage overhead, but also serious risk of privacy breach due to the centralized entity suffering from single point of failure \cite{dibconsortium}. To solve these challenges, an emerging distributed machine learning technique named federated learning is introduced to allow mobile devices to jointly train a shared  global model  in a decentralized manner. The mobile devices only  send local model updates  trained on their local raw data to a task publisher of federated learning  without uploading any raw data, thus decoupling  the machine learning from acquiring, storing and training data in a central server \cite{federated2019}. 

With the significant advantages in privacy protection,  federated learning has attracted increasing attention from researchers and developers recently.
Google designed a virtual keyboard application named Gboard for smart phones by using  federated learning \cite{hard2018federated}.  The authors in \cite{yang2019federated} further discussed  architecture and potential applications about federated learning. The authors in \cite{federated2019} formulated an optimization problem of federated learning over wireless networks to obtain optimal learning time, accuracy level, and energy cost. A deep Q learning algorithm is used  to solve the optimal data and energy management problems of federated learning  without prior knowledge of network dynamics in  \cite{anh2018efficient}. Considering clients with heterogeneous resources, the authors in \cite{nishio2018client} proposed a client selection scheme  for federated learning based on a greedy algorithm.


The aforementioned studies have specifically focused on optimizing the performance of federated learning algorithms, e.g., learning time or energy cost. However, the most existing work made an optimistic assumption that all the mobile devices will unconditionally participate in federated learning when invited  \cite{shayan2018biscotti,kim2018device}, which is not practical in the real world due to resource costs incurred by model training   \cite{zhou2019computation}. Without  well-designed economic compensation, the self-interested  mobile devices will be reluctant to participate in  federated learning \cite{federated2019,kim2018device}.
Moreover,  there exist the following information asymmetry issues between the task publisher and the mobile devices. 
 I) The task publisher does not know the amount of available  computation resources and the data sizes from  mobile devices for model training.  II) The  local data quality of a  mobile device is  unknown to the task publisher due to the lack of prior knowledge. As a result, the task publisher may incur a high cost when providing  incentives to the mobile devices. Therefore, it is essential for the task publishers to design an efficient incentive mechanism to reduce the impact of information asymmetry \cite{yang2019federated,anh2018efficient}.


In this paper, to attract mobile devices with high-quality data  to join federated learning and overcome the information asymmetry issue, we adopt the contract theory to  design an efficient incentive mechanism that maps the contributed resources into appropriate rewards. The data owners  (i.e., mobile devices) with larger-accuracy and more reliable  local data and more resource contributions  can obtain more rewards from the task publisher. Each data owner chooses its desired contract item to maximize its profit \cite{zhou2019computation}.
The main contributions of this paper are listed as follows:
\begin{itemize}
	\item We design an effective incentive mechanism using  contract theory to stimulate  mobile devices to join  federated model training under information asymmetry.
	
	\item To attract data owners with high-quality data, we define the quality-related parameter of local data as the  type of the contract model. The higher type data owners that have larger-accuracy and more reliable  local data can receive more rewards.
	
	\item We perform the real-world experiments using the well-known digit classification dataset to demonstrate that the proposed mechanism outperforms  existing approaches. 
\end{itemize}

The rest of this paper is organized as follows. The problem formulation and solutions for contract theory model are introduced in Section \ref{problem} and Section \ref{solution}, respectively. Numerical results are presented in Section \ref{simulation} followed by the conclusions in Section \ref{conclusion}.

\section{Problem Formulation} \label{problem}
Similar to that in \cite{federated2019},  we consider a universal mobile network with  wireless communication infrastructures  and a set of mobile devices. These widely distributed communication infrastructures, e.g., base stations, can act as task publishers with federated learning tasks, while the mobile devices are  data owner candidates for the learning tasks. 
The mobile devices equipped  with advanced computation and communication functions can not only generate diverse user data from mobile applications, but also collect a lot of sensing data.   Each task publisher designs contract items for incentivizing mobile devices with high-quality data to join federated learning. Every data owner iteratively trains a  shared global model with local model updates generated using their private local data. Then, all the data owners upload their local model updates to the task publisher for updating the global model. The training process is repeated until the accuracy of the global model achieves a pre-defined, desirable value. More details about the basics of federated learning can be found in \cite{federated2019,zhu2018multi}.

\subsection{Computation Model for Federated Learning}
We consider a federated learning task as a monopoly market with a monopolist operator (a task publisher) and a set of mobile devices $\mathcal{N}=\{1,\dots,N\}$.  Each data owner $n \in \mathcal{N}$ with a local training dataset uses a size $s_n$ of its local data samples  to participate in the federated learning task. There is an input-output pair  in each data sample, in which the input is a sample vector with various data features and the output is the label value for the input generated through mobile apps \cite{federated2019}.   The contributed computation resources for local model training, i.e., CPU cycle frequency, from the data owner $n$ is denoted as  $f_n$. The number of CPU cycles for a data owner $n$ to perform  local model training using a single data sample \footnote{We consider that each data sample has the same data size.} is denoted by $c_n$. Hence, for data owner $n$, the computation time of a local iteration in local model training is $\frac{{{c_n}{s_n}}}{{{f_n}}}$.  According to \cite{federated2019}, the CPU energy consumption  for one local iteration is   $E_n^{cmp}({f_n}) = \zeta {c_n}{s_n}f_n^2,$
where  $\zeta$ is the effective capacitance parameter of  the computing chipset for data owner $n$.

\subsection{Communication Model for Federated Learning}
For a federated learning task,  all the participating  data owners  collaborate to train a shared global model and achieve a global accuracy level of learning by an iterative method with a number of communication rounds (i.e., global iterations).  During a global iteration, the data owners send their own local model updates to the task publisher through wireless communications.
 \textbf{Each local model update from worker $n$ is affected by its local data quality, which is denoted as  $\varepsilon _n$. The local data quality $\varepsilon _n$   mainly depends on local data accuracy and data reliability, and can be normalized to a range. Note that, more accurate or reliable data brings larger $\varepsilon _n$. Intuitively,   a better data quality (i.e., larger value of $\varepsilon _n$) leads to fewer local  and  global iterations and also improves the accuracy of training models \cite{dataquality2019}. For ease of analysis,  we use  ${\log (\frac{1 }{{{\varepsilon _n}}})}$  to represent the number of iterations of a local model update  when the global accuracy is fixed \cite{federated2019,konevcny2016federated}, which can be easily extended to more complicated expressions.}
The computation time of a local iteration and uplink communication time\footnote{We consider that the downlink time between the task publisher and the data owners is negligible compared with the uplink time as typically  the downlink bandwidth is much larger than the uplink bandwidth.} of a local model update are involved in  a global iteration. The computation time of a local iteration by data owner $n$ is denoted by $T_n^{cmp}=\frac{{{c_n}{s_n}}}{{{f_n}}}$.
For the communication time of local model updates, time-sharing multi-access protocols, e.g., Time-Division Medium Access (TDMA) technology, are taken into consideration in this paper.  We consider that the locations of data owners are fixed when transmitting local model parameters. The transmission rate of data owner $n$ is denoted as ${r_n} = B{\ln}(1 + \frac{{{{\rho _n}}{h_n}}}{{{N_0}}})$  \cite{federated2019}.  Here, $B$ is the transmission bandwidth and ${\rho _n}$ is the transmission power of the data owner $n$. $h_n$ is the channel gain of peer-to-peer link between  data owner $n$ and the task publisher.  $N_0$ is the background noise.  We consider the data size of a local model update  $\sigma $ to be a constant with the same value for all data owners.
The transmission time of a local model update with size $\sigma $ is expressed by $T_n^{com} = \frac{{{\sigma }}}{{B{\ln}(1 + \frac{{{\rho _n}{h_n}}}{{{N_0}}})}}$.

Therefore, the total time of participating in one global iteration for the data owner $n$ is denoted as 
\begin{equation}
{T_n^t} = \log (\frac{1}{{{\varepsilon _n}}})T_n^{cmp} + T_n^{com} .
\end{equation}


According to \cite{federated2019}, the energy consumption by data owner $n$ to transmit local model updates in a global iteration  is expressed as $E_n^{com} = T_n^{com} \cdot {{\rho _n}} = \frac{{\sigma {{\rho _n}}}}{{B\ln (1 + \frac{{{{{\rho _n}}}{h_n}}}{{{N_0}}})}}$.
Therefore, for a global iteration, the total energy consumption of the data owner $n$ is denoted as follows:
\begin{equation}
E_n^t =\log (\frac{1}{{{\varepsilon _n}}})E_n^{cmp}+ E_n^{com}.
\end{equation}

\subsection{Profit Function of the Task Publisher}
To attract more data owners with high-quality data (i.e., high-accuracy and reliable local data), we define data quality as the type of a data owner $n$, which is denoted as ${\theta _n} = \frac{\psi }{{\log (\frac{1}{{{\varepsilon _n}}})}}.$
Here, $\psi$ is the coefficient about the number of local model iterations affected by the local data  quality. The data owners are divided into $M$ types sorted in ascending order of data quality: ${\theta _1}<\cdots<{\theta _m}<\cdots<{\theta _M}, m\in\{1,\dots,M\}.$ A larger ${\theta _m}$ means  better data quality with  higher accuracy and reliability leading to fewer local model iterations \cite{nishio2018client,kang2018towards}.
Although the task publisher does not know exactly true type of a given data owner, it has the knowledge of the probability that a data owner belongs to a certain type-\textit{m} \cite{zhang2016offloading}  and $\sum\nolimits_{m = 1}^M {{p_m}}  = 1$. The task publisher obtains the distribution of data owner types from previous observations \cite{zeng2018icc}.

Due to information asymmetry, the task publisher should design specific contracts for different types of data owners with different levels of data quality to increase  its profits. The task publisher offers different  resource-reward bundles to  the data owners according to their types. For different data owners with different computation resources, i.e., CPU cycle frequency, the task publisher offers  the contract   $(R_n(f_n), f_n)$ including a series of resource-reward bundles. Here,  $f_n$ is the computation resource of type-\textit{n} a data owner and $R_n(f_n)$ is the corresponding reward for the data owner. The more contributed computation resource leads to faster local model training, thus bringing  higher rewards.  The data owners choose and sign one of the provided contracts at will and finish the given federated learning task. If  a data owner cannot finish the learning task or misbehaves, the task publisher will  put the data owner into a blacklist and withhold payment.

For a signed contract  $(R_n(f_n), f_n)$,  we define the profit of the task publisher obtained from a type-\textit{n} data owner as ${U_{TP}}({R_n}) = \omega \ln ({T_{\max }} - T_n^t) - l{R_n},$
where $\omega >0$ is the  satisfaction degree parameter of task publisher. $T_{max}$ is the task publisher's maximum tolerance time of  federated learning, and   $l$ is the unit cost about the rewards for the data owners.  $[\omega \ln ({T_{\max }} - T_n^t)]$ is the satisfaction function of the task publisher regarding the total time of one global iteration  for type-\textit{n} data owner.  Note that both the higher quality (higher type) and  larger CPU cycle frequency can improve the profit for the task publisher, i.e.,   $\frac{{\partial {U_{TP}}}}{{\partial {\varepsilon _n}}} > 0$, $\frac{{\partial {U_{TP}}}}{{\partial {\theta_n}}} > 0$ and $\frac{{\partial {U_{TP}}}}{{\partial {f_n}}} > 0$. Moreover, for the task publisher,  more high-type data owners joining the federated learning lead to more profit, but also incur larger reward cost $lR_n$. Apparently, the task publisher will not accept a negative profit when performing the federated learning task, i.e., ${U_{TP}}({R_n}) \ge 0$.
The objective of the task publisher is to maximize its profit in the federated learning task defined as follows:
 \begin{equation}
\begin{array}{l}
\mathop {\max }\limits_{({{\rm{R}}_{\rm{n}}},{{\rm{f}}_{\rm{n}}})} {U_{TP}} = \sum\limits_{n = 1}^N {N{p_n} \cdot } w\ln [{T_{\max }} - (\frac{\sigma }{{B\ln (1 + \frac{{{p_n}{h_n}}}{{{N_0}}})}}+\\
~~~~~~~~~~~~~~~~~~ \frac{\psi}{{{\theta _n}}} \cdot \frac{{{c_n}{s_n}}}{{{f_n}}})] - l{R_n}.
\end{array} \label{equ1}
 \end{equation}

\subsection{Utility Function of Data Owners}
The utility function of a type-\textit{n} data owner for the signed contract $(R_n(f_n), f_n)$ is defined as: ${U_D}({f_n}) = {R_n} - \mu  E_n^t=  {R_n} - \mu \left[\frac{{\psi}}{{{\theta _n}}}\zeta {c_n}{s_n}f_n^2 + E_n^{com}\right],$
where $\mu$ is a pre-defined weight parameter for energy consumption. We consider that every data owner is self-interested and  the valuation of ${U_D}$ is zero when there is no reward \cite{zhang2016offloading}. Intuitively,  the higher-type data owners have larger utility since they provide better  quality data. The data owner also wishes to minimize energy consumption when performing the federated learning  task for maximizing its utility. The overall goal of a type-\textit{n} data owner is expressed by
\begin{eqnarray}
\mathop {\max }\limits_{({{\rm{R}}_{\rm{n}}},{{\rm{f}}_{\rm{n}}})} {U_D} = {R_n} - \mu \left[ \frac{\psi}{\theta _n}\zeta {c_n}{s_n}f_n^2 + \frac{{\sigma {{\rho _n}}}}{{B\ln (1 + \frac{{{{\rho _n}}{h_n}}}{{{N_0}}})}}\right]. \label{equ2}
\end{eqnarray}

\section{Optimal Contract Designing}\label{solution}
With information asymmetry, to make contracts feasible, each contract must satisfy the following  constraints: i) Individual Rationality (IR) and ii) Incentive Compatibility (IC) in order to ensure that each type of data owners are properly motivated \cite{zhang2016offloading}.

\textbf{Definition 1.} {Individual Rationality}:  A data owner only participates in the federated learning task when its utility  is not less than zero, i.e.,
\begin{equation}
{U_D} = {R_n} - \mu \left[\frac{\psi}{\theta _n}\zeta {c_n}{s_n}f_n^2 + \frac{{\sigma {{\rho _n}}}}{{B\ln (1 + \frac{{{{\rho _n}}{h_n}}}{{{N_0}}})}}\right] \ge 0.
\end{equation}

\textbf{Definition 2.} {Incentive Compatibility}: To maximize utility, every data owner can only choose the contract designed for itself, i.e., type $\theta_n$  instead of any other contracts $(R_{m}, f_{m})$, i.e.,
 \begin{equation}
\begin{array}{l}
{R_n} - \mu [\frac{\psi}{\theta _n}\zeta {c_n}{s_n}f_n^2 + \frac{{\sigma {{\rho _n}}}}{{B\ln (1 + \frac{{{{\rho _n}}{h_n}}}{{{N_0}}})}}] \ge {R_{m}} -  \mu [\frac{\psi}{{{\theta _{m}}}}\zeta {c_n}{s_n}f_n^2 \\  + \frac{{\sigma {{\rho _n}}}}{{B\ln (1 + \frac{{{{\rho _n}}{h_n}}}{{{N_0}}})}}],\forall n,m \in \{ 1,\ldots,N\}, n \ne m.
\end{array}
 \end{equation}

In what follows,  for simplicity, we consider $\mu=1$. Without loss of generality, we consider the transmission bandwidth, transmission power, and the channel gain for all the data owners to be identical  due to similar wireless communication environments \cite{federated2019}, and thus we have $E_1^{com} =  \cdots  = E_n^{com} = \frac{{\sigma {\rho _0}}}{{B\ln (1 + \frac{{{\rho _0}{h_0}}}{{{N_0}}})}},n \in \{ 1, \ldots ,N\}$.  For ease of presentation, the optimization problems in (\ref{equ1}) and (\ref{equ2})  can be reformulated as
\begin{equation}
\begin{array}{*{20}{l}}
{\mathop {{\rm{max}}}\limits_{{\rm{(}}{{\rm{R}}_{\rm{n}}}{\rm{,}}{{\rm{f}}_{\rm{n}}}{\rm{)}}} {U_{TP}} = \sum\limits_{n = 1}^N {N {p_n}} \left[w\ln ({T_{\max }} - T_n^{com} - \frac{\psi{T_n^{cmp}}}{{{\theta _n}}}) - l{R_n}\right]},\\
{{\rm{s}}.{\rm{t}}.\;}\\
{{R_n} - (\frac{\psi}{\theta _n}E_n^{cmp} + E_n^{com}) \ge 0,\forall n \in \{ 1, \ldots ,N\} ,}\\
{{R_n} - (\frac{\psi}{\theta _n}E_n^{cmp} + E_n^{com}) \ge {R_m} - (\frac{\psi}{\theta _n}E_m^{cmp} + E_m^{com}),}\\
{\forall n,m \in \{ 1,\ldots,N\} ,n \ne m,}\\
{\frac{{{c_n}{s_n}}}{{{f_n}}} \le {T_{\max }},\forall n \in \left\{ {1, \ldots ,N} \right\},}\\
{\sum\limits_{n = 1}^N {N \cdot {p_n}}  \cdot {R_n} \le {R_{\max }},\forall n \in \left\{ {1, \ldots ,N} \right\},}
\end{array}\label{5}
\end{equation}
where $R_{max}$ is the total reward budget of the task publisher.
Although the problem in (\ref{5}) is not a convex optimization problem, its solution can be found  by performing the following transformation.

According to the above definitions, we have the following lemmas.

\textbf{Lemma 1 (Monotonicity).} For contract $(R_n, f_n)$ and $(R_m, f_m)$, we have   $f_n \geq f_m$ and $R_n \geq R_m$, if and only if $\theta_n \geq \theta_m$, $n \ne m$, and $n, m \in\{1,\dots,N\}$.

\textbf{Lemma 2.} If the IR constraint of type-\textit{1} is satisfied, the other IR constraints will also hold.


\textbf{Lemma 3.} According to the monotonicity in \textbf{Lemma 1}, the IC condition can be reduced as the Local Downward Incentive Constraints (LDIC) that is expressed as ${R_n} - \frac{\psi}{\theta _n}E_n^{cmp} \ge {R_{n - 1}} - \frac{\psi}{\theta _n}E_{n - 1}^{cmp},\forall n \in \{ 2,\dots,N\}.$

The proofs of \textbf{Lemma 1}, \textbf{2}, and \textbf{3} are similar to those in the \cite{kang2018towards}.
Based on the analysis of these lemmas, the optimization problem in (\ref{5}) is simplified as follows:

\begin{equation}
\begin{array}{l}
\mathop {{\rm{max}}}\limits_{{\rm{(}}{{\rm{R}}_{\rm{n}}},{{\rm{f}}_{\rm{n}}}{\rm{)}}} {U_{TP}} = \sum\limits_{n = 1}^N {N  {p_n}}\left[w\ln ({T_{\max }} - T_n^{com} - \frac{\psi{T_n^{cmp}}}{{{\theta _n}}}) - l{R_n}\right],\\
{\rm{s}}.{\rm{t}}.\;\\
{R_n} - \frac{\psi}{\theta _n}E_n^{cmp} - E_n^{com} = 0,\forall n \in \{ 1,...,N\} ,\\
{{R_n} - \frac{\psi}{\theta _n}E_n^{cmp} = {R_{n - 1}} - \frac{\psi}{\theta _n}E_{n - 1}^{cmp},\forall n \in \{ 2,...,N\} ,}\\
{\frac{{{c_n}{s_n}}}{{{f_n}}} \le {T_{\max }},\forall n \in \left\{ {1, \ldots ,N} \right\},}\\
\sum\limits_{n = 1}^N {N \cdot {p_n}}  \cdot {R_n} \le {R_{\max }},\forall n \in \left\{ {1, \ldots ,N} \right\}.
\end{array}
\label{28}
\end{equation}

To derive the optimal contracts in the problem (\ref{28}), we first solve the relaxed problem  in (\ref{28}) without monotonicity constraint. Subsequently, this acquired solution is checked whether it satisfies the monotonicity condition.  By using the iterative method on \textit{IC} and \textit{IR} constraints, we  can obtain the reward which is expressed as ${R_n} = E_n^{com} + \frac{\psi{E_1^{cmp}}}{{{\theta _1}}} + \sum\limits_{k = 1}^n {{\Delta _k}},$
where ${\Delta _k} = \frac{\psi{E_k^{cmp}}}{{{\theta _k}}} - \frac{\psi{E_{k - 1}^{cmp}}}{{{\theta _k}}}$ and ${\Delta _1} = 0$.  By substituting $R_n$ into $\sum\limits_{n = 1}^N {N \cdot {p_n}}  \cdot {lR_n}$, we can obtain
\begin{equation}
\sum\limits_{n = 1}^N {N \cdot {p_n}}  \cdot l{R_n} =NlE_n^{com} + Nl\zeta \sum\limits_{n = 1}^N {{g_n}} {c_n}{s_n}f_n^2,
\label{30}
\end{equation}
where 
${g_n} = \left\{ {\begin{array}{*{20}{l}}
{\frac{\psi{{P_n}}}{{{\theta _n}}} + (\frac{\psi}{\theta _n} - \frac{1}{{{\theta _{n + 1}}}})\sum\limits_{i = n + 1}^N {{p_i},~n < N,} }\\
{\frac{\psi{{P_N}}}{{{\theta _N}}},~~~~~~~~~~~~~~~~~~~~~~~~~~~~~n = N.}
	\end{array}} \right.$

By substituting (\ref{30}) into the problem  in (\ref{28}) and also removing all $R_n$,  we can rewrite (\ref{28}) as
\begin{equation}
\begin{array}{*{20}{l}}
{\mathop {{\rm{max}}}\limits_{({{\rm{R}}_{\rm{n}}},{{\rm{f}}_{\rm{n}}})} {U_{TP}} = \sum\limits_{n = 1}^N {N{p_n}} \left[ {w\ln ({T_{\max }} - \frac{\sigma }{{B\ln (1 + \frac{{{\rho _n}{h_n}}}{{{N_n}}})}} - \frac{{\psi {c_n}{s_n}}}{{{f_n}{\theta _n}}})} \right]}\\
{\;\;\;\;\;\;\;\;\;\;\;\;\;\;\;\;\; - \frac{{Nl\sigma {\rho _n}}}{{B\ln (1 + \frac{{{\rho _n}{h_n}}}{{{N_n}}})}} - Nl\zeta \sum\limits_{n = 1}^N {{c_n}{s_n}{g_n}} f_n^2,}\\
{{\rm{s}}.{\rm{t}}.}\\
{\frac{{{c_n}{s_n}}}{{{T_{\max }}}} \le {f_n},\forall n \in \left\{ {1, \ldots ,N} \right\},}\\
{\frac{{N\sigma {\rho _n}}}{{B\ln (1 + \frac{{{\rho _n}{h_n}}}{{{N_n}}})}} + N\zeta \sum\limits_{n = 1}^N {{c_n}{s_n}{g_n}} f_n^2 \le {R_{\max }},\forall n \in \left\{ {1, \ldots ,N} \right\}.}
\end{array} \label{33}
\end{equation}

By differentiating $U_{TP}$ with respect to $f_n$, we can obtain  $\frac{{{\partial ^2}{U_{TP}}}}{{\partial f_n^2}} < 0$, and thus  $U_{TP}$ is concave.  The summation of concave functions  ($U_{TP}$) is still a concave function,  and hence the problem in (\ref{33}) with affine constraints is a concave optimization problem. With the help of convex optimization tools, e.g., CVX, we can calculate the optimal computation resource, i.e., contributed CPU-cycle $f_n^*$ and the corresponding incentive $R_n^*$ \cite{kang2018towards}.  In addition,  the monotonicity can be automatically met when the types of data owners follow uniform distribution. If the distribution of  data owners' types is not uniform, we can utilize the infeasible sub-sequence replacing algorithm to meet the final optimal computation resource requirement \cite{kang2018towards,hou2018incentive}.

\section{Numerical Results} \label{simulation}
In the simulation,  a well-known digit classification dataset named MNIST is used  to evaluate the performance of the proposed incentive schemes. This dataset includes 60,000 training examples and 10,000 test examples, which can be used to perform a digit classification task.  We  consider a task publisher and 100 data owners in the federated learning tasks. The data owners are randomly assigned a training set following a uniform distribution over 10 classes as their own local training data. The accuracy of local data ranges from 20\% to 92\%. The CPU cycles of performing a data sample $c_n$ is 5 and the size of data samples $s_n$ is 20. The transmission time $T_n^{com}$ and energy consumption $E_n^{com}$ for transmitting a local model update are 10 and 20, respectively. The maximum tolerance time $T_{max}$ and the total amount of given reward $R_{max}$ of a federated learning task are 600 and 10,000, respectively.  Moreover, the data owners are initially classified into 10 types according to quality-related parameters of local training data, and the probability for a candidate belonging to a certain type is 0.1 \cite{kang2018towards}. 
\begin{figure}[t]\centering
	\includegraphics[width=0.35\textwidth]{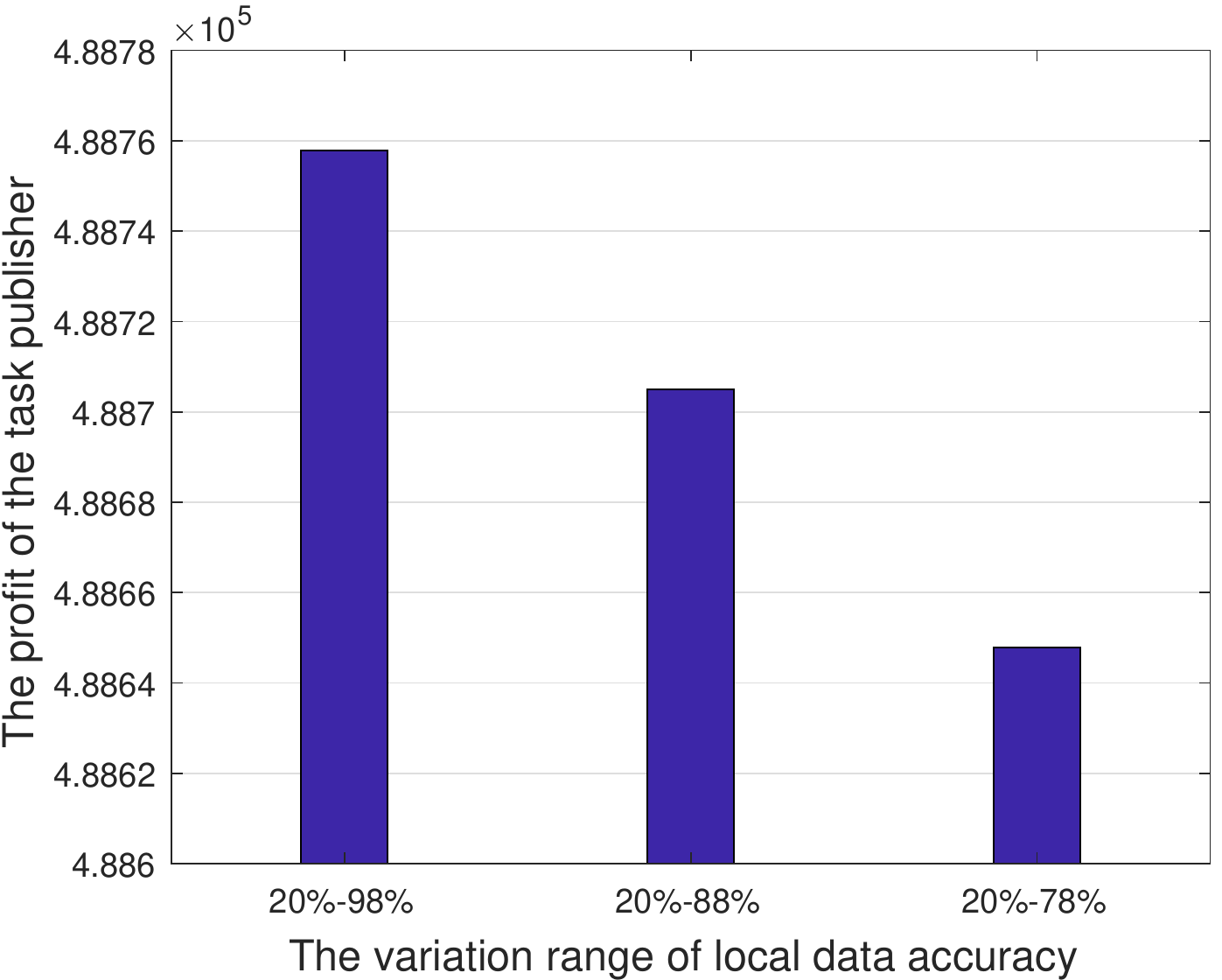}
	\caption{The profit of the task publisher with respect to different accuracy levels of local training data.}
	\label{SW2Accu1} 
\end{figure}

\begin{figure}[t]\centering
	\includegraphics[width=0.35\textwidth]{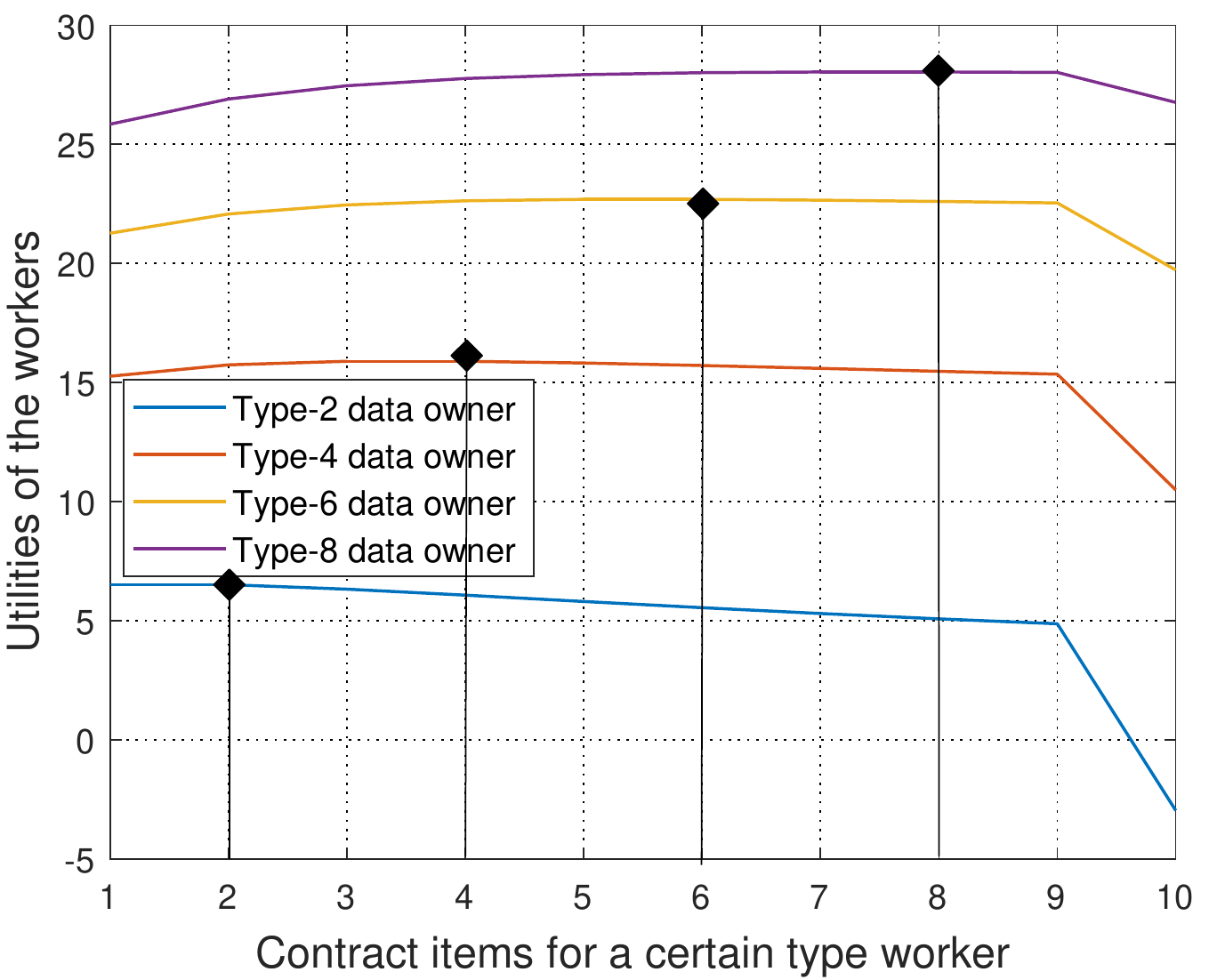}
	\caption{Utilities of data owners with different contract items.}
	\label{typeWorkerUnity} \vspace*{-3mm}
\end{figure}

\begin{figure}[t]\centering
	\includegraphics[width=0.35\textwidth]{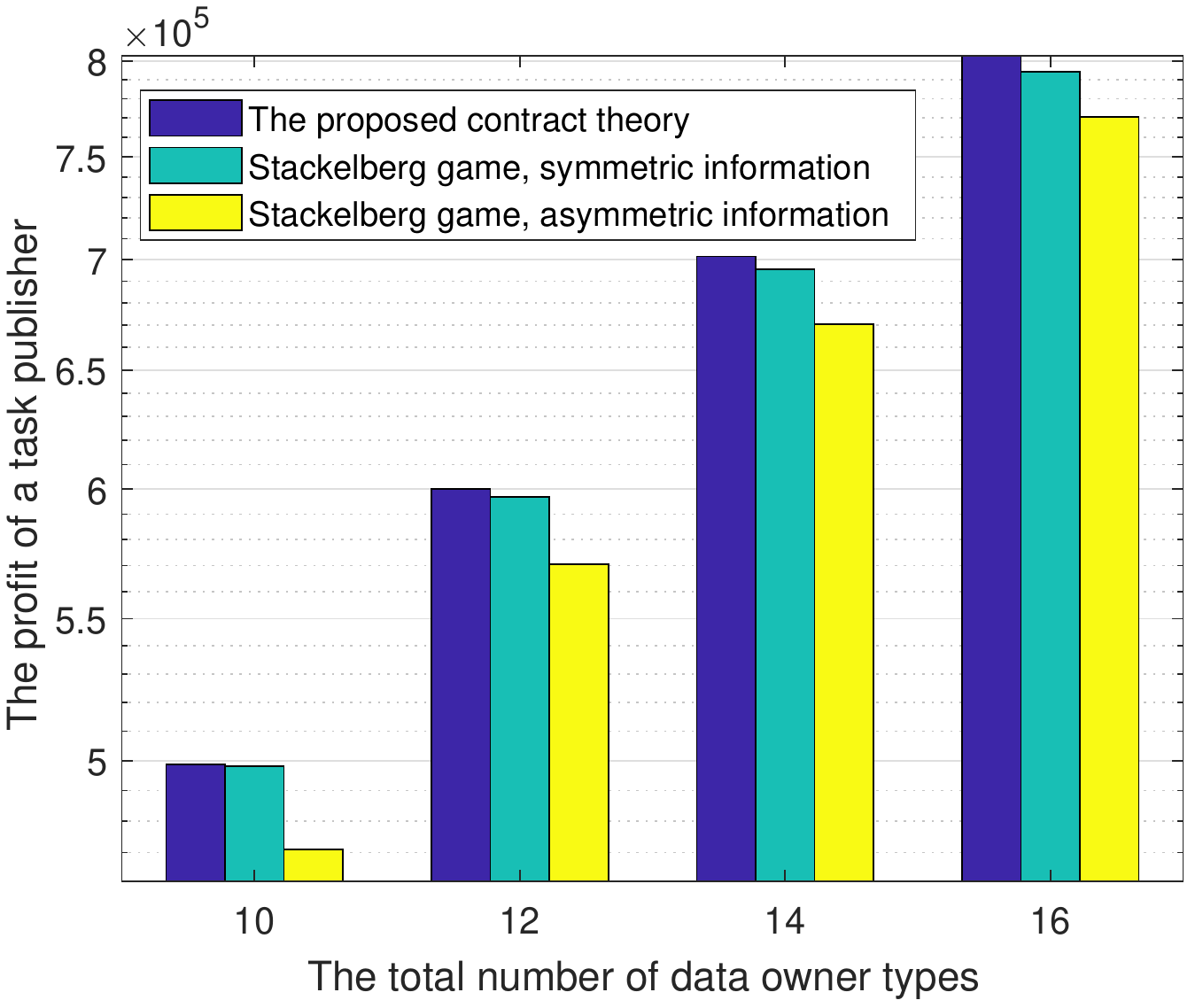}
	\caption{The profit of a task publisher with respect to different total number of data owner types}
	\label{taskPulbisherUtility}  \vspace*{-4mm}
\end{figure}


To show the impacts of the variation range of local training data accuracy on the profit of task publisher, we vary the upper limit of local data accuracy (i.e., a quality parameter related to the data owner type) from 98\% to 78\%, respectively. As shown in Fig.~\ref{SW2Accu1}, the profit of the task publisher decreases with the decrease of the upper limit of local data accuracy. As reducing upper limit of the local data accuracy means that the number of high-type data owners is decreasing. Therefore, the low-quality of local training data has a negative impact on the profit of the task publisher. 
Therefore, the proposed schemes can stimulate data owners with high-quality data  to join learning tasks, hence leading to  more efficient federated learning.

To validate the feasibility, i.e., IR and IC, of the proposed scheme under information asymmetry, we present Fig.~\ref{typeWorkerUnity} to show the utilities of data owners with types 2, 4, 6 and 8, respectively \cite{hou2018incentive}. From Fig.~\ref{typeWorkerUnity}, we observe that all types of data owners can only achieve their own maximum utility when they choose the contract item exactly designed for their types, which explains the IC constraint \cite{kang2018towards}. Moreover, each data owner can obtain nonnegative utility when selecting the contract item corresponding to its type, which validates the IR constraint. 

We compare the profit of the task publisher obtained from the proposed contract theory model, and that from the Stackelberg game model in ~\cite{hou2018incentive}.  Figure~\ref{taskPulbisherUtility} shows that the larger total number of data owner types leads to the larger profit of a task publisher. The more data owner types bring more contract item choices to high-type data owners, thus ensuring more efficient federated learning. For a certain number of data owner types,  the profit of the task publisher in  the proposed contract model is higher than that of the Stackelberg game model \cite{kang2018towards}. The reason is that, in the monopoly market, the task publisher working as the monopolist only provides limited contract items to the data owners and extracts more profit from the data owners. Nevertheless, in the Stackelberg game model, rational  data owners can optimize their individual utilities resulting in  less profit for the task publisher. Although the task publisher needs to consider the IR and IC constraints during designing the contract items,  these constraints have a small impact on maximizing the utilities of the data owners compared with the Stackelberg game model \cite{liu2017design}. As a result, the task publisher can obtain the higher profit than that in the Stackelberg game models \cite{kang2018towards}.
Moreover, the Stackelberg game model with symmetric information has better performance than that of Stackelberg game model with asymmetric information. The reason is that the game leader (the task publisher) in the Stackelberg game with symmetric information can optimize its profit because of knowledge about the actions of  followers (data owners), i.e., the symmetric information, and set the utilities of the followers to zero \cite{hou2018incentive}.


\section{Conclusions} \label{conclusion}
In this paper, we designed a contract theory-based incentive mechanism  to motivate  data owners that have high-quality local training data to join the learning processes for efficient  federated learning. Numerical results have indicated that the proposed incentive scheme can attract more  data owners with high-quality local training data to ensure efficient federated learning and also optimize the utilities of both the task publishers and the data owners. For further work, we will consider using blockchains to ensure reliability of local model updates   when formulating the incentive mechanism for reliable federated learning in mobile networks \cite{kangwcl,zhangdrl,iotj2018,liu2018blockchain}. 

%
%
%


\begin{thebibliography}{10}
	
	\bibitem{yang2019federated}
	Q.~Yang \textit{et al.,} ``Federated machine learning: Concept
	and applications,'' {\em ACM Transactions on Intelligent Systems and
		Technology (TIST)}, vol.~10, no.~2, pp.~12:1--12:15, 2019.
	
	\bibitem{zhu2018multi}
	H.~Zhu and Y.~Jin, ``Multi-objective evolutionary federated learning,'' {\em arXiv preprint arXiv:1812.07478}, 2018.
	
	\bibitem{dibconsortium}
	X.~Zhu, H.~Li, and Y.~Yu, ``Blockchain-based privacy preserving deep
	learning,'' 2018. Available:
	\url{http://xxhb.fjnu.edu.cn/_upload/tpl/06/5d/1629/template1629/papers/83.pdf}.
	
	\bibitem{federated2019}
	H.~T. Nguyen \textit{et al.,} ``Federated
	learning over wireless networks: Optimization model design and analysis,'' in
	{\em IEEE INFOCOM 2019}, Paris, France, April 2019.
	
	\bibitem{hard2018federated}
	A.~Hard \textit{et al.,} ``Federated learning for mobile keyboard prediction,'' {\em
		arXiv preprint arXiv:1811.03604}, 2018.
	
	\bibitem{anh2018efficient}
	T.~T. Anh \textit{et al.,} ``Efficient
	training management for mobile crowd-machine learning: A deep reinforcement
	learning approach,'' {\em arXiv preprint arXiv:1812.03633}, 2018.
	
	\bibitem{nishio2018client}
	T.~Nishio and R.~Yonetani, ``Client selection for federated learning with
	heterogeneous resources in mobile edge,'' {\em arXiv preprint
		arXiv:1804.08333}, 2018.
	
	\bibitem{shayan2018biscotti}
	M.~Shayan \textit{et al.,}  ``Biscotti: A ledger for
	private and secure peer-to-peer machine learning,'' {\em arXiv preprint
		arXiv:1811.09904}, 2018.
	
	\bibitem{kim2018device}
     H. Kim \textit{et al.,} ``Blockchained On-Device Federated Learning,''  \textit{IEEE Communications Letters}, in press, 2019.
	
	\bibitem{zhou2019computation}
	Z.~Zhou \textit{et al.,}  ``Computation	resource allocation and task assignment optimization in vehicular fog
	computing: A contract-matching approach,'' {\em IEEE Transactions on Vehicular Technology}, in press, 2019.
	
	  \bibitem{dataquality2019}
	A.~Richardson~\emph{et~al.},	``Rewarding High-Quality Data via Influence Functions,'' 2019. [online] Available: \url{https://arxiv.org/abs/1908.11598}	
	
	\bibitem{konevcny2016federated}
	J.~Kone{\v{c}}n{\`y} \textit{et al.,} 	``Federated optimization: Distributed machine learning for on-device
	intelligence,'' {\em arXiv preprint arXiv:1610.02527}, 2016.
	
	\bibitem{kang2018towards}
	J.~{Kang} \textit{et al.,} 
	``Toward secure blockchain-enabled internet of vehicles: Optimizing consensus
	management using reputation and contract theory,'' {\em IEEE Transactions on
		Vehicular Technology}, vol.~68, pp.~2906--2920, March 2019.
	
	\bibitem{zhang2016offloading}
	Y.~Zhang \textit{et al.,} ``Offloading in
	software defined network at edge with information asymmetry: A contract
	theoretical approach,'' {\em Journal of Signal Processing Systems}, vol.~83,
	no.~2, pp.~241--253, 2016.
	
	\bibitem{zeng2018icc}
	M.~Zeng \textit{et al.,} ``Incentive mechanism design
	for computation offloading in heterogeneous fog computing: A contract-based
	approach,''  {\em IEEE International Conference on Communications
		(ICC)}, pp.~1--6, IEEE, 2018.
	
	\bibitem{hou2018incentive}
	Z.~Hou, H.~Chen, Y.~Li, and B.~Vucetic, ``Incentive mechanism design for
	wireless energy harvesting-based internet of things,'' {\em IEEE Internet of
		Things Journal}, vol.~5, no.~4, pp.~2620--2632, 2018.
	
	\bibitem{liu2017design}
	T.~Liu \textit{et al.,}  ``Design of contract-based
	trading mechanism for a small-cell caching system,'' {\em IEEE Transactions
		on Wireless Communications}, vol.~16, no.~10, pp.~6602--6617, 2017.
	
	\bibitem{kangwcl}
	J.~{Kang} \textit{et al.,} 
	``Incentivizing consensus propagation in proof-of-stake based consortium
	blockchain networks,'' {\em IEEE Wireless Communications Letters}, vol.~8,
	pp.~157--160, Feb 2019.
	
	\bibitem{zhangdrl}
	Y.~Dai  \textit{et al.,}  ``Blockchain and deep
	reinforcement learning empowered intelligent 5g beyond,'' {\em IEEE Network
		Magazine}, vol.~33, no.~3, 2019.
	
	\bibitem{iotj2018}
	J.~{Kang} \textit{et al.,} ``Blockchain for secure and efficient data sharing in vehicular
	edge computing and networks,'' {\em IEEE Internet of Things Journal}, vol.~6,
	pp.~4660--4670, June 2019.
	
	\bibitem{liu2018blockchain}
	H.~Liu, Y.~Zhang, and T.~Yang, ``Blockchain-enabled security in electric
	vehicles cloud and edge computing,'' {\em IEEE Network}, vol.~32, no.~3,
	pp.~78--83, 2018.
	
\end{thebibliography}
\end{document}